\documentclass[10pt,twocolumn,letterpaper]{article}

\usepackage{cvpr}              % To produce the CAMERA-READY version
\usepackage{algorithm}
\usepackage{algpseudocode}
\algrenewcommand\algorithmicrequire{\textbf{Input:}}
\algrenewcommand\algorithmicensure{\textbf{Output:}}
\algrenewcommand\textproc{\textsc}

\usepackage{amsmath}
\usepackage{amssymb}
\usepackage{booktabs}
\usepackage{tikz}
\usepackage{cuted}

\definecolor{cvprblue}{rgb}{0.21,0.49,0.74}
\usepackage[pagebackref,breaklinks,colorlinks,allcolors=cvprblue]{hyperref}
\usepackage{multirow}
\usepackage{marvosym}
\usepackage[table]{xcolor}
\newcommand{\CC}[1]{\cellcolor{gray!#1}}

\title{Towards Holistic Modeling for Video Frame Interpolation with \\ Auto-regressive Diffusion Transformers}

\author{Xinyu Peng$^{1,\dag}$,
Han Li$^{1,\dag}$,
Yuyang Huang$^{1}$,
Ziyang Zheng$^{1(\textrm{\Letter})}$,
Yaoming Wang$^{1(\textrm{\Letter})}$,\\
Xin Chen$^{2}$,
Wenrui Dai$^{1}$,
Chenglin Li$^{1}$,
Junni Zou$^{1}$, and
Hongkai Xiong$^{1}$\\
$^1$Shanghai Jiao Tong University $^2$Fudan University.\\
{\tt\small \{xypeng9903, qingshi9974, huangyuyang, zhengziyang, wang\_yaoming, daiwenrui,}\\
{\tt\small lcl1985, zoujunni, xionghongkai\}@sjtu.edu.cn}
, {\tt\small chenxin061@gmail.com}
}

\begin{document}
\maketitle

\newcommand\blfootnote[1]{% 
\begingroup 
\renewcommand\thefootnote{}\footnote{#1}% 
\addtocounter{footnote}{-1}% 
\endgroup 
}
\blfootnote{$^\dag$ Indicates equal contribution.}
\blfootnote{$^{\textrm{\Letter}}$ Correspondence to: Ziyang Zheng \textless zhengziyang@sjtu.edu.cn\textgreater, Yaoming Wang \textless wang\_yaoming@sjtu.edu.cn\textgreater.}

% \begin{abstract}
% % Video frame interpolation is a challenging task, particularly for long video sequences where temporal coherence is crucial. 
% Existing methods for video frame interpolation (VFI) often process video in short, independent segments, e.g., triplets, leading to temporal inconsistencies and motion artifacts. To address this, we propose \textbf{L}ocal \textbf{D}iffusion \textbf{F}orcing Transformers for \textbf{V}ideo \textbf{F}rame \textbf{I}nterpolation (LDF-VFI), which adopts a novel, holistic, and video-centric VFI paradigm. Our framework leverages an auto-regressive diffusion transformer equipped with sparse, local attention to efficiently capture long-range dependencies while maintaining efficient memory usage. Benefiting from this local design, our model can generalize to arbitrary spatial resolutions at inference without retraining when coupled with tiled VAE encoding, enabling supporting high-resolution videos (e.g., 4K) without prohibitive computational cost. To mitigate error accumulation during auto-regressive inference, we introduce a skip-concatenate sampling strategy to ensure temporal stability. We further employ an enhanced conditional VAE decoder that incorporates multi-scale features from the input low-frame-rate video, thereby improving reconstruction fidelity. Empirically, LDF-VFI achieves state-of-the-art performance, particularly on challenging long sequences with large motion, outperforming existing methods in both per-frame quality and temporal consistency.
% \end{abstract}

\begin{abstract}
% Video frame interpolation is a challenging task, particularly for long video sequences where temporal coherence is crucial. 
Existing video frame interpolation (VFI) methods often adopt a frame-centric approach, processing videos as independent short segments (e.g., triplets), which leads to temporal inconsistencies and motion artifacts. To overcome this, we propose a holistic, video-centric paradigm named \textbf{L}ocal \textbf{D}iffusion \textbf{F}orcing for \textbf{V}ideo \textbf{F}rame \textbf{I}nterpolation (LDF-VFI). Our framework is built upon an auto-regressive diffusion transformer that models the entire video sequence to ensure long-range temporal coherence. To mitigate error accumulation inherent in auto-regressive generation, we introduce a novel skip-concatenate sampling strategy that effectively maintains temporal stability. Furthermore, LDF-VFI incorporates sparse, local attention and tiled VAE encoding, a combination that not only enables efficient processing of long sequences but also allows generalization to arbitrary spatial resolutions (e.g., 4K) at inference without retraining. An enhanced conditional VAE decoder, which leverages multi-scale features from the input video, further improves reconstruction fidelity. Empirically, LDF-VFI achieves state-of-the-art performance on challenging VFI benchmarks, demonstrating superior per-frame quality and temporal consistency, especially in scenes with large motion. The source code is available at \url{https://github.com/xypeng9903/LDF-VFI}.

\end{abstract}

\section{Introduction}\label{sec:intro}

\begin{figure}
    \centering
    \includegraphics[width=1\linewidth]{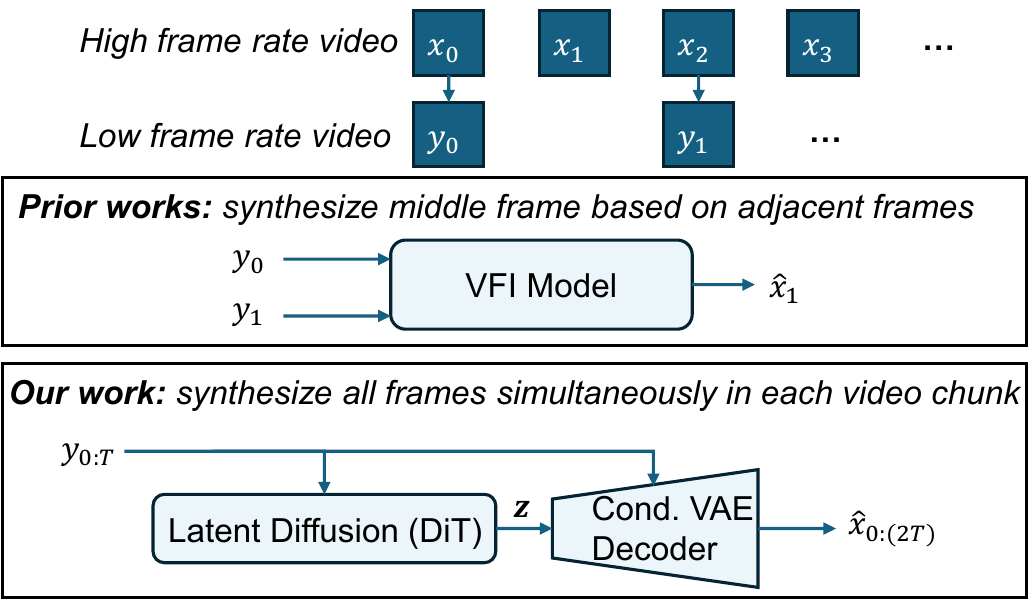}
    \caption{ Comparison between prior works and our framework. We illustrate $2\times$ VFI for simplicity. Prior methods treat VFI as a set of independent triplets, ignoring dependencies among interpolated frames. In contrast, our framework models the entire video holistically, leading to improved temporal coherence.}
    \label{fig:compare-modeling}
    \vspace{-10pt}
\end{figure}

Video frame interpolation (VFI), a classic computer vision task, aims to synthesize intermediate frames between existing ones and is essential for applications like slow-motion~\cite{huang2022rife,jain2024video}, novel view synthesis~\cite{chen2022videoinr}, and video generation~\cite{team2025longcat,singer2022make}. However, conventional VFI methods that rely on optical flow~\cite{huang2022rife,li2023amt,zhang2023emavfi,seo2025bim} often struggle with complex, non-linear motions, leading to a recent shift towards diffusion models~\cite{danier2024ldmvfi,zhang2025eden,lbbdm,madiff}.

While promising, these methods suffer from two primary limitations. \textbf{First}, as illustrated in \Cref{fig:compare-modeling}, they typically operate on independent frame triplets (i.e., synthesizing a middle frame $\widehat{x}_1$ from adjacent input frames $y_0$ and $y_1$), while ignoring the temporal dependencies between subsequently interpolated frames (e.g., $\{\widehat{x}_1, \widehat{x}_2,\widehat{x}_3,\cdots\}$). This \textit{frame-centric} approach can lead to significant temporal inconsistencies and motion artifacts. \textbf{Second}, using vanilla diffusion transformer (DiT) architecture~\cite{peebles2023scalable} faces prohibitive computational costs, as the quadratic complexity of full attention renders them infeasible for supporting high-resolution videos (e.g., 4K).

An ideal approach for maintaining VFI temporal consistency would be to synthesize all frames for the entire video in a single inference. However, this is computationally intractable. To overcome these limitations, we introduce a new VFI paradigm designed to: 1) decompose the input video into multiple temporal chunks, and synthesize all frames for each chunk in a single inference to ensure intra-chunk consistency; 2) connect the synthesized frames of all chunks auto-regressively to maintain long-range inter-chunk consistency; and 3) scale efficiently to high resolutions such as 4K videos.

To achieve the objectives, we propose a \textit{video-centric} framework, \textbf{L}ocal \textbf{D}iffusion \textbf{F}orcing for \textbf{V}ideo \textbf{F}rame \textbf{I}nterpolation (LDF-VFI), built upon an auto-regressive diffusion transformer. LDF-VFI first processes video in fixed-length temporal chunks, enabling processing arbitrarily long sequences with constant memory per step. In each chunk, all frames are synthesized in a single inference, ensuring intra-chunk consistency. 

To address the challenge of inter-chunk modeling, we employ auto-regressive generation for VFI. A key challenge in this paradigm is error accumulation, also known as \textit{exposure bias}~\cite{schmidt2019generalization}. This phenomenon occurs when the model conditions on its own imperfect outputs, causing errors to propagate and degrade quality over time. We mitigate this by introducing a novel \textit{skip-concatenate sampling} strategy. Unlike traditional causal inference, where each newly generated segment depends on the immediately preceding (and potentially imperfect) output, our method periodically generates a chunk independently of the most recent context. This \textit{skip} chunk effectively resets the accumulated error. A subsequent \textit{concatenate} chunk is then generated, conditioned on both the new \textit{skip} chunk and the prior context, thereby seamlessly bridging the temporal gap. This strategy ensures temporal stability over long sequences by preventing uncontrolled error propagation. 

To efficiently achieve resolution scalability, we adopt a sparse, local attention scheme, which is spatially local but temporally dense. This design avoids the quadratic complexity of full attention and, combined with tiled VAE encoding, enabling seamless generalization to resolutions unseen during training. Furthermore, to enhance the visual quality and fidelity of the generated frames, we design a \textit{conditional VAE decoder}. Standard VAEs often lose fine-grained details during reconstruction from the latent space. Our enhanced decoder mitigates this by incorporating multi-scale features extracted from the input low-frame-rate video. By injecting this rich contextual information throughout the decoding process, our model produces sharper and more accurate reconstructions that are faithful to the original scene.

To validate the effectiveness of our video-centric design, we develop \textit{video-centric benchmarks}, SNU-FILM-entire and X4K-entire, which evaluate interpolation quality on full videos rather than isolated frame triplets. Extensive experiments show that LDF-VFI achieves state-of-the-art performance, with clear gains in both per-frame perceptual quality and long-range temporal coherence, particularly under large and complex motion.

\section{Related work}

\subsection{Video frame interpolation} 
% Most recent video frame interpolation (VFI) methods estimate intermediate optical flow and synthesize the target frame via forward warping~[40] or backward warping~[13, 19, 28, 30, 34]. For example, SGM-VFI~[31] uses sparse global matching to refine flow, while VFIMamba~[66] adopts a global receptive field via state-space models to capture motion. Diffusion-based approaches have also emerged: LDMVFI~[14] employs a latent diffusion model conditioned on the input frames, LBBDM~[36] applies consecutive Brownian-bridge diffusion, MADIFF~[21] conditions diffusion on motion cues, and VIDIM~[23] cascades diffusion models for target-frame generation. Despite their progress, diffusion-based methods can still struggle with large, complex motions and maintaining long-range temporal coherence. LDF-VFI addresses these issues with video-centric modeling, stronger latent representations, and an auto-regressive inference scheme that reduces exposure bias.

Video frame interpolation (VFI) methods fall into two major categories: optical-flow-based (and kernel-based) and diffusion-based approaches. Flow-based methods~\cite{huang2022rife,li2023amt,zhang2023emavfi,seo2025bim} estimate intermediate motion and warp input frames forward or backward, while kernel-based methods predict adaptive local kernels to aggregate pixels from input frames for interpolation~\cite{niklaus2017video,niklaus2018context,lee2020adacof}. Recent flow-based variants such as SGM-VFI~\cite{sgmvfi} and VFIMamba~\cite{vfimamba} enhance robustness through sparse global matching or global receptive fields. These approaches work well for simple and approximately linear motions, but their performance degrades significantly in complex scenarios with non-linear dynamics, occlusions, or large displacements. Diffusion-based VFI has recently emerged, including latent-diffusion-based LDMVFI~\cite{danier2024ldmvfi} and EDEN~\cite{zhang2025eden}, Brownian-bridge diffusion in LBBDM~\cite{lbbdm}, motion-conditioned diffusion in MADIFF~\cite{madiff}, and cascaded diffusion models in VIDIM~\cite{jain2024video}. 
% Although diffusion models better handle complex and ambiguous motions, they tend to introduce hallucinations, require heavy iterative computation, and face a fundamental trade-off between efficiency and temporal coherence. 
Recent diffusion-based methods interpolate frames pairwise, which eases computation but often causes temporal inconsistency under challenging large motion scenarios. These limitations make existing VFI solutions difficult to apply in real-world settings.

% {\color{blue}
% \subsection{Video diffusion models}

% Recent advances in image and video generation have demonstrated the effectiveness of diffusion models for high-dimensional, continuous data. Video diffusion models extend spatial denoising to the temporal dimension, leveraging 3D architectures or factorized spatio-temporal operators, as well as conditioning mechanisms tailored to video tasks.

% Our work follows this line but focuses specifically on interpolation with strong conditioning from observed frames and a design that supports long sequences.
% }

\subsection{Auto-regressive generative modeling}
Auto-regressive (AR) models represent a fundamental approach to generative modeling, factorizing the joint probability distribution of a sequence into a product of conditionals via the chain rule: $p(x) = \prod_{i=1}^{N} p(x_i | x_{<i})$. This principle has been widely applied across various domains, from text models~\cite{ranzato2015sequence, brown2020language} to high-dimensional data such as images~\cite{chen2020generative, esser2021taming, VAR, Infinity} and videos~\cite{yu2023language, InfinityStar}, as well as large-scale AR modeling across various data modalities~\cite{wang2024emu3,li2025synergen,chen2025janus,li2025onecat, cui2025emu3,peng2025hyperet}.

More recently, a promising direction has emerged in hybrid models that integrate auto-regressive frameworks with diffusion models, particularly for tasks like video generation~\cite{chen2024diffusion, li2025arlon, zhang2025test, zhang2025generative, huang2025self}. These approaches typically operate by auto-regressively generating a sequence of data segments (e.g., frames), where each new segment is conditioned on previously generated ones. Despite their power, these methods inherit a fundamental challenge of AR modeling known as \textit{exposure bias}~\cite{schmidt2019generalization}. This issue arises from the discrepancy between training, which uses ground-truth context, and inference, which conditions on the model's own imperfect outputs. This mismatch can lead to an accumulation of errors, progressively degrading the quality of long sequences. Various strategies have been proposed to mitigate this problem—such as scheduled sampling~\cite{bengio2015scheduled}, optimizing alternative objectives like reverse KL-divergence~\cite{huang2025self}, or directly training the model on corrupted condition~\cite{Infinity,ruhe2024rolling,xie2025progressive}. In this paper, we propose a VFI-specific solution, called skip-concatenate sampling, to mitigate exposure bias and achieve temporal consistency for arbitrary long generation.

% While these diffusion-forcing works primarily focus on unconditional or text-to-video generation \textcolor{blue}{[TODO: cite diffusion forcing A, B]}, we introduce a novel skip-concatenate sampling strategy specifically designed for VFI. Benefiting from strong input conditioning, our method keeps error accumulation constant, avoiding the temporal drift seen in open-ended generation. Furthermore, unlike training-based self-correction approaches \textcolor{blue}{[TODO: cite exposure bias solution C]} which degrade on videos exceeding training lengths, our strategy mitigates exposure bias during inference, smoothly enabling robust arbitrary-length generation.

\section{Methodology}

\subsection{Preliminaries}

\paragraph{Flow matching}
Flow matching~\cite{lipman2022flow}, closely related to diffusion models~\cite{ho2020denoising, song2020score,hong2026improving}, is popular for modeling high dimensional distributions by learning the velocity field that transports a simple prior distribution (e.g., Gaussian noise) to a complex data distribution. Given a data sample $\mathbf{x}_0 \sim q_0(\mathbf{x})$ and a noise sample $\mathbf{x}_1 \sim q_1(\mathbf{x})$, we define a linear interpolation path $\mathbf{x}_t = (1-t)\mathbf{x}_0 + t\mathbf{x}_1$ for $t \in [0,1]$. Denote the distribution of $\mathbf{x}_t$ is $q_t$, the time derivative of $q_t$ can be described by \textit{the continuity equation} $\partial_t q_t(\mathbf{x}) = -\nabla\cdot(q_t(\mathbf{x})\, \mathbf{v}_t(\mathbf{x}))$. Therefore, the sample evolving according to the ODE: $\mathrm{d}\mathbf{x}_t =\mathbf{v}_t(\mathbf{x}_t) \mathrm{d}t$ also follows $\mathbf{x}_t \sim q_t$. Remarkably, a simple L2 regression objective $\mathcal{L}_{\text{FM}}$ admits the true velocity field $\mathbf{v}_t(\mathbf{x}_t)$ as its global minimizer that is leveraged to train the flow matching model $\mathbf{v}_\theta(\mathbf{x}_t, t)$:
\begin{equation}\label{eq:fm-loss}
\mathcal{L}_{\text{FM}} = \mathbb{E}_{t, q_0(\mathbf{x}_0) q_1(\mathbf{x}_1)} \big[ \| \mathbf{v}_\theta(\mathbf{x}_t, t) - (\mathbf{x}_1 - \mathbf{x}_0) \|^2 \big].
\end{equation}
At inference time, starting from $\mathbf{x}_1 \sim q_1$, we simulate the ODE $\mathrm{d}\mathbf{x}_t = \mathbf{v}_\theta(\mathbf{x}_t,t) \mathrm{d}t$ from $t=1$ to $0$ using numerical ODE solver to obtain an approximate data sample $\mathbf{x}_0$.

\paragraph{Tiled VAE encoding}
Tiled VAE encoding is a technique to scale VAE encoding to high resolutions while avoiding seams at tiles' boundaries. Each frame is split into tiles of size $(H_{\min}, W_{\min})$ with stride $(s_H, s_W)$. Every tile is encoded independently by the VAE, then overlaps are blended in latent space with a simple linear ramp: rows from the tile above and columns from the tile to the left are mixed proportionally across the overlap width/height. After blending, only the stride region of each latent tile (the non-overlapped interior) is kept and arranged into a compact latent. Decoding inverts the procedure: latent tiles are decoded, then blended in pixel space with the same vertical and horizontal ramps, and their stride regions are concatenated to form a seam‑free reconstruction. This yields (1) constant memory cost, (2) smooth boundaries, and (3) a regular latent grid that downstream sparse attention can exploit.~\footnote{For practical considerations, please refer to the open-source implementation: \url{https://github.com/huggingface/diffusers/blob/main/src/diffusers/models/autoencoders/autoencoder_kl_wan.py}}
% \enlargethispage{-0.8\baselineskip}

\subsection{Holistic modeling for VFI}
We revisit the fundamental problem of video frame interpolation (VFI). Prevailing approaches typically frame VFI as an independent prediction task, where an intermediate frame $\hat{I}_{t}$ is synthesized from two adjacent frames, $I_0$ and $I_1$~\cite{huang2022rife,guo2024generalizable,zhang2023emavfi,jin2023uprnet,li2023amt}. This frame-centric formulation, while direct, presents two significant limitations. First, by treating each interpolation in isolation, it overlooks the crucial temporal correlations required for motion consistency across a sequence of newly generated frames. Second, it disregards the broader temporal context available in non-adjacent frames (e.g., $I_{-1}, I_2$).

% {\color{blue}Consequently, these methods often introduce temporal artifacts and unnatural motion, particularly when handing video with large, nonlinear motion.}

\begin{figure*}
    \centering
    \includegraphics[width=1\linewidth]{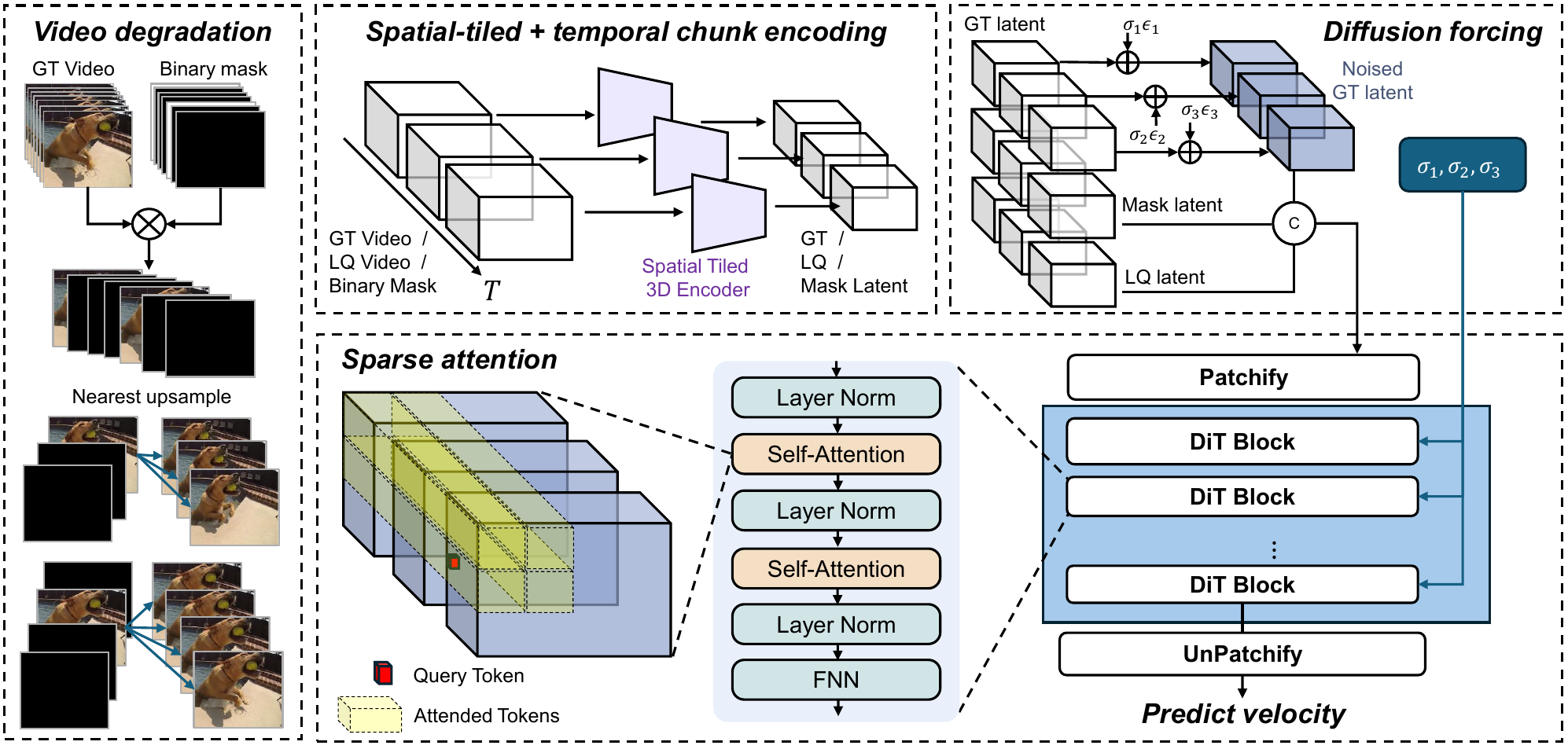}
    \caption{Model architecture overview. Our model first aligns the low-quality (LQ) video with the high-quality (HQ) ground-truth (GT) video's temporal dimension using nearest-neighbor upsampling. Both the GT video and the upsampled LQ video, along with a corresponding binary mask, are then processed through a spatial-tiled and temporal non-overlapping chunk encoder to generate latent representations. During training, we employ diffusion forcing on these temporal chunks, a strategy that enables auto-regressive inference. Furthermore, the model incorporates sparse attention to efficiently handle high-resolution video inputs, ensuring scalability.}
    \label{fig:arch}
\end{figure*}

\begin{figure}
    \centering
    \includegraphics[width=1.0\linewidth]{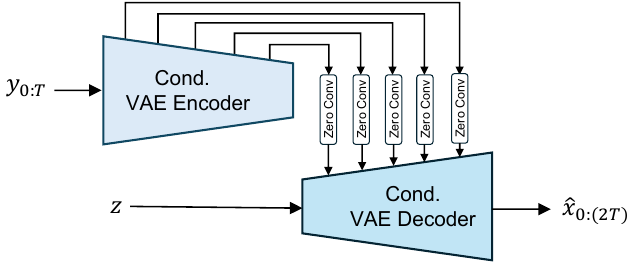}
    \caption{The architecture of conditional VAE decoder.}
    \label{fig:cond-vae}
    \vspace{-10pt}
\end{figure}

\paragraph{Modeling entire video sequence}
To overcome these limitations, we advocate for a \textit{video-centric perspective}. Instead of predicting individual frames, we model the joint conditional probability distribution of the \textit{entire high-frame-rate video sequence} conditioned on the \textit{entire sparse input frames}. Formally, given a low-frame-rate video $\mathbf{y}\in \mathbb{R}^{T \times H \times W \times 3}$ of $T$ frames, our objective is to learn the full conditional distribution $q(\mathbf{x} \mid \mathbf{y})$ of the corresponding high-frame-rate video $\mathbf{x}\in \mathbb{R}^{(sT) \times H \times W \times 3}$, where $s$ is the temporal upsampling factor. By modeling the entire sequence holistically, this approach ensures the generated intermediate frames are both visually plausible and temporally coherent, preserving the natural dynamics of the video.

\paragraph{Latent space generation}
Motivated by the success of latent diffusion for modeling high-dimensional continuous data~\cite{rombach2022high,brooks2024video,wan2025}, we introduce a latent variable $\mathbf{z}\in \mathbb{R}^{T' \times H' \times W' \times C}$ to perform generative modeling in a compact, computationally efficient latent space. Specifically, we extend $q(\mathbf{x} \mid \mathbf{y})$ to $q(\mathbf{x}, \mathbf{z} \mid \mathbf{y})$ and define a generative model $p_{\theta}(\mathbf{x}, \mathbf{z} \mid \mathbf{y})$ learned to fit $q(\mathbf{x},\mathbf{z} \mid \mathbf{y})$ as follows:
\begin{align}
    &q(\mathbf{x}, \mathbf{z} \mid \mathbf{y}) := \underbrace{q(\mathbf{x} \mid \mathbf{y})}_{\text{Target}}\underbrace{q(\mathbf{z} \mid \mathbf{x})}_{\text{VAE encoder}} \\
    &p_{\theta}(\mathbf{x}, \mathbf{z} \mid \mathbf{y}) := \underbrace{p_{\theta}(\mathbf{z} \mid \mathbf{y})}_{\text{Latent diffusion}}\underbrace{p_{\theta}( \mathbf{x} \mid \mathbf{z}, \mathbf{y})}_{\text{VAE decoder}}
\end{align}
Here, the joint distribution $q(\mathbf{x}, \mathbf{z} \mid \mathbf{y})$ is defined by our target distribution $q(\mathbf{x} \mid \mathbf{y})$ and an off-the-shelf VAE encoder $q(\mathbf{z} \mid \mathbf{x})$. Our generative VFI model, $p_{\theta}(\mathbf{x}, \mathbf{z} \mid \mathbf{y})$, comprises two key components: i) a latent diffusion (or flow-matching) model $p_{\theta}(\mathbf{z} \mid \mathbf{y})$ that fits the marginal latent distribution $q(\mathbf{z} \mid \mathbf{y})=\int q(\mathbf{x}, \mathbf{z} \mid \mathbf{y}) \, \mathrm{d}\mathbf{x}$, and ii) a conditional VAE decoder $p_{\theta}(\mathbf{x} \mid \mathbf{z}, \mathbf{y})$ that reconstructs the final high-frame-rate video from latents $\mathbf{z}$, conditioned on the input low-frame-rate video $\mathbf{y}$. We detail our architectural choices in Section~\ref{sec:arch}, including the DiT backbone for $p_{\theta}(\mathbf{z} \mid \mathbf{y})$, and a modified VAE decoder for $p_{\theta}(\mathbf{x} \mid \mathbf{z}, \mathbf{y})$.

\begin{figure*}[!t]
    \centering
    \includegraphics[width=1\linewidth]{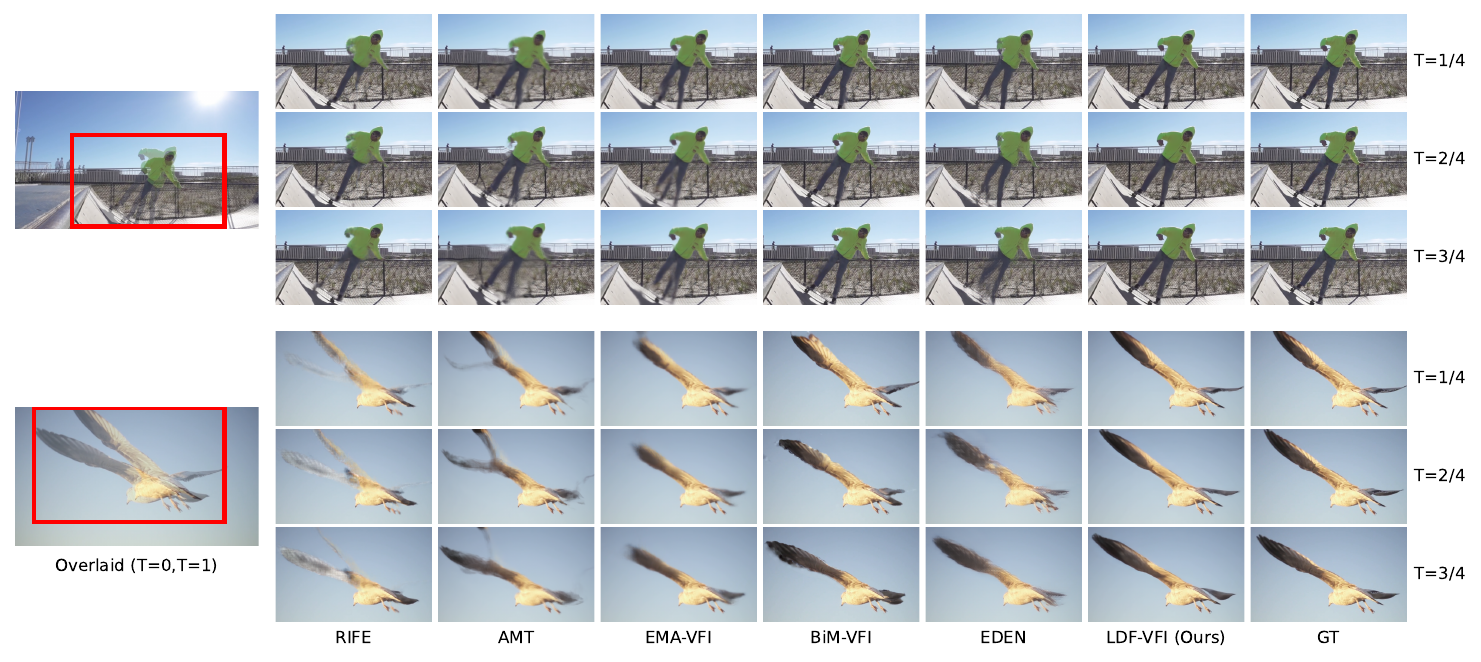}
    \caption{Qualitative comparisons of $16\times$ interpolation on SNU-FILM-entire benchmark.}
    \label{fig:compare-baseline}
\end{figure*}

\begin{figure}[!t]
    \centering
    \includegraphics[width=1\linewidth]{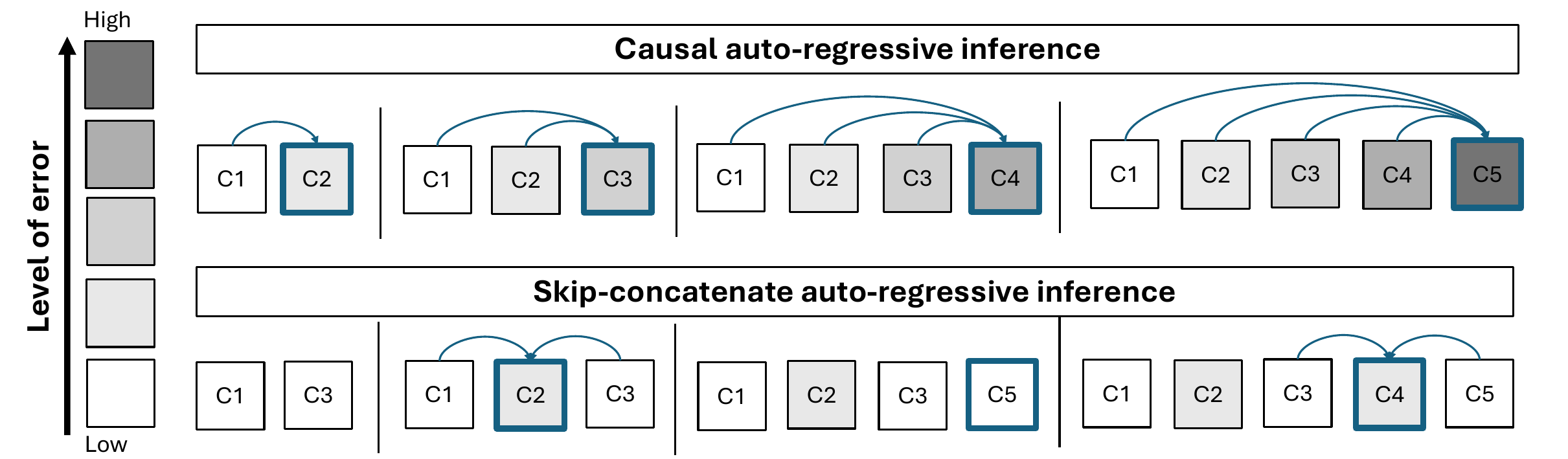}
    \caption{Causal vs. skip-concatenate auto-regressive inference. A bold blue border marks the current chunk being generated, and a blue arrow indicates its context. Darker chunks signify greater error accumulation.}
    \label{fig:concatenate-vs-causal}
    \vspace{-10pt}
\end{figure}

\subsection{Model architecture}\label{sec:arch}
We modified a 3D DiT backbone from Wan2.1~\cite{wan2025} with local spatio-temporal attention, optimized for latent-space generation under strong video conditioning. Specifically, our modifications include: 1) restricting attention to be local in space but global in time, which enables high-resolution processing while capturing complex motion; 2) introducing nearest-neighbor temporal upsampling for improved LQ information preservation; and 3) injecting LQ video embeddings into the original VAE decoder to better guide the VAE reconstruction. An overview of our model architecture is presented in Figure~\ref{fig:arch}.

\paragraph{Video conditioning} Recent studies have explored multiple strategies for conditioning DiT models on video inputs, including channel concatenation~\cite{wan2025}, cross-attention~\cite{ye2025stylemaster}, and in-context conditioning~\cite{cai2025videocanvas}, among others~\cite{hacohen2024ltx}. We adopt channel concatenation for its simplicity and the explicit spatio-temporal correspondence it preserves between low-frame-rate input and target high-frame-rate sequences, e.g., missing frames can be represented via zero padding~\cite{wan2025}. However, prior work~\cite{cai2025videocanvas} has demonstrated that off-the-shelf pretrained VAEs are not robust to such irregularly padded inputs, necessitating VAE retraining to address the induced distribution shift. To this end, we propose a simple alternative: \textit{nearest-neighbor temporal upsampling} combined with a binary mask that identifies positions corresponding to true observed frames. We encode the LQ video using the same VAE encoder as for HQ video, whereas the binary mask is encoded through simple nearest spatial downsampling and a pixel shuffle (time-to-channels) operator for temporal downsampling. By chunking this properly precomputed binary mask consistently with the output video and assigning the corresponding segment to each chunk, our model naturally extends to VFI at arbitrary and non-integer downsampling ratios. 

% We empirically demonstrate that these approaches preserve more LQ information than zero padding in Section~\ref{sec:abla}.

\paragraph{Local VAE encoding} To enable our method to scale to high-resolution videos, we adapt a \textit{spatial tiled VAE encoding} scheme that processes the video in overlapping spatial patches, significantly reducing memory overhead. In contrast to the spatial domain, we employ a \textit{temporal non-overlapping} strategy. The video is partitioned into fixed-length temporal chunks, and each chunk is encoded independently by the VAE. This design provides discrete latent units that are essential for our auto-regressive inference framework, as will be detailed in Section~\ref{sec:ar-training}. Empirically, we find that this non-overlap encoding introduces no discernible temporal incoherence, and the DiT model is able to learn the temporal continuity between the generated chunks.

\paragraph{Sparse attention mechanism}
To manage the computational demands of high-resolution video, we employ a sparse attention strategy within our DiT that leverages the native locality from the tiled VAE encoding. The latent video is naturally partitioned into a grid of spatio-temporal chunks, for which we design a hybrid sparse attention mechanism. For spatial dimensions, we implement a chunk-based sliding window attention. This approach confines attention within local chunks and allows the receptive field to grow efficiently without the quadratic computational cost of full attention. For the temporal dimension, we apply full attention across all chunks within the processing window. Since the temporal length is fixed and relatively short, this enables the model to capture complex, non-local temporal correlations without computational bottlenecks. This hybrid strategy—sparse in space but dense in time—is tailored for video processing, enabling the model to learn intricate motion patterns while remaining scalable.

\begin{table*}[!t]
    \caption{Quantitative results on SNU-FILM-entire benchmarks. We use \textbf{bold} and \underline{underline} to highlight the best and second-best results.}
    \centering \small
    \setlength{\tabcolsep}{1.1pt}
    % \resizebox{\linewidth}{!}{
    \begin{tabular}{l cccc cccc cccc}
\toprule
        \multirow{2}{*}{Method} & \multicolumn{4}{c}{SNU-FILM-4$\times$} & \multicolumn{4}{c}{SNU-FILM-8$\times$} & \multicolumn{4}{c}{SNU-FILM-16$\times$} \\
        \cmidrule(lr){2-5} \cmidrule(lr){6-9} \cmidrule(lr){10-13}
        & LPIPS$\downarrow$ & FVD$\downarrow$ & VFIPS$\downarrow$ & FloLPIPS$\downarrow$ 
        & LPIPS$\downarrow$ & FVD$\downarrow$ & VFIPS$\downarrow$ & FloLPIPS$\downarrow$ 
        & LPIPS$\downarrow$ & FVD$\downarrow$ & VFIPS$\downarrow$ & FloLPIPS$\downarrow$ \\
\midrule
        RIFE~\cite{huang2022rife} [ECCV'22]     
        & \underline{0.028} & \underline{9.02} & \textbf{-90.14} & 0.047 
        & 0.054 & 25.30 & \underline{-83.11} & 0.098 
        & 0.110 & 69.76  & -65.41 & 0.182 \\
        
        AMT~\cite{li2023amt} [CVPR'23]      
        & 0.125 & 21.02 & -73.08 & 0.147 
        & 0.230 & 93.31 & -43.55 & 0.283 
        & 0.360 & 263.60 & -4.74  & 0.419 \\
        
        EMA-VFI~\cite{zhang2023emavfi} [CVPR'23]  
        & 0.050 & 15.28 & -86.33 & 0.069 
        & 0.083 & 38.16 & -78.46 & 0.121 
        & 0.115 & 101.67 & -62.93 & 0.209 \\
        
        BiM-VFI~\cite{seo2025bim} [CVPR'25]  
        & 0.032 & 10.78 & -88.60 & 0.044 
        & 0.046 & \underline{21.52} & -62.92 & 0.209 
        & \textbf{0.074} & \underline{38.26}  & \textbf{-72.31} & \underline{0.118} \\
        
        EDEN~\cite{zhang2025eden} [CVPR'25]     
        & \textbf{0.021} & 10.64 & \underline{-89.16} & \textbf{0.036} 
        & \textbf{0.040} & 26.03 & -82.66 & \underline{0.070} 
        & 0.078 & 58.83 & -68.13 & 0.128 \\
\midrule
        \textbf{LDF-VFI \textit{(ours)}} 
        & \CC{15}0.032 & \CC{15}\textbf{8.21} & \CC{15}-87.14 & \CC{15}\underline{0.042} 
        & \CC{15}\underline{0.045} & \CC{15}\textbf{15.01} & \CC{15}\textbf{-83.78} & \CC{15}\textbf{0.063} 
        & \CC{15}\underline{0.078} & \CC{15}\textbf{26.26} & \CC{15}\underline{-71.48} & \CC{15}\textbf{0.117} \\
\bottomrule
    \end{tabular} 
    % }
    \label{tab:main}
\end{table*}

\begin{table}[!t]
    \caption{Quantitative results on X4K-entire benchmark. We use \textbf{bold} and \underline{underline} to highlight the best and second-best results.}
    \centering
    \setlength{\tabcolsep}{4pt}
    \resizebox{\columnwidth}{!}{
    \begin{tabular}{l cccc}
\toprule
        \multirow{2}{*}{Method} & \multicolumn{4}{c}{X4K-16$\times$} \\
        \cmidrule(lr){2-5}
        & LPIPS$\downarrow$ & FVD$\downarrow$ & VFIPS$\downarrow$ & FloLPIPS$\downarrow$ \\
\midrule
        RIFE~\cite{huang2022rife} [ECCV'2022]     
        & 0.176 & 328.56 & -28.88 & 0.201 \\
        
        AMT~\cite{li2023amt} [CVPR'2023]      
        & 0.123 & 157.29 & -55.71 & 0.151 \\
        
        EMA-VFI~\cite{zhang2023emavfi} [CVPR'2023]  
        & 0.109 & 76.67 & -63.91 & 0.158 \\
        
        BiM-VFI~\cite{seo2025bim} [CVPR'2025]  
        & \textbf{0.055} & \underline{69.83} & \textbf{-69.98} & \textbf{0.065} \\
        
        EDEN~\cite{zhang2025eden} [CVPR'2025]     
        & 0.454 & 1586.99 & 21.23 & 0.491 \\
\midrule
        \textbf{LDF-VFI \textit{(ours)}} 
        & \CC{15}\underline{0.071} & \CC{15}\textbf{51.41} & \CC{15}\underline{-64.80} & \CC{15}\underline{0.082} \\
\bottomrule
    \end{tabular} 
    }
    \label{tab:x4k}
\end{table}

\paragraph{Conditional VAE}
To enhance reconstruction quality, we introduce a \textit{conditional VAE decoder} inspired by ControlNet~\cite{zhang2023adding}. As illustrated in Figure~\ref{fig:cond-vae}, this architecture uses a dedicated conditional encoder, which mirrors the main decoder's structure, to extract multi-scale spatio-temporal features from the low-frame-rate input video. These features are then injected into the corresponding layers of the main decoder via zero-initialized convolutional layers and residual connections. This multi-scale conditioning strategy provides fine-grained guidance throughout the reconstruction process, leveraging both global and local details from the input video to significantly improve the temporal coherence and visual quality of the final output.

\subsection{Auto-regressive inference with diffusion forcing} \label{sec:ar-training}
While modeling the full video sequence is ideal, the unbounded length of real-world videos makes it impractical to process the entire sequence at once due to prohibitive memory and computational requirements. To address this scalability challenge, we adopt an \textit{temporal auto-regressive approach} that enables handling videos of arbitrary length with constant memory and computational cost per step, regardless of the total duration, as elaborated below.

\paragraph{Diffusion forcing training} 
We adopt \textit{diffusion forcing}~\cite{chen2024diffusion} at the temporal chunk level to train the DiT. Specifically, we partition the ground-truth video, the upsampled low-frame-rate input, and the corresponding binary mask into a sequence of fixed-length temporal chunks. Instead of applying a single noise level to the entire sequence, we sample an \textit{independent} noise level for each encoded VAE latent of ground-truth chunk. The model then learns to predict the velocity conditioned on other chunks at \textit{arbitrary noise levels}. This allows the model to predict the velocity conditioned on cleaned, previous generated chunks, enabling auto-regressive sampling at the inference time.

% filepath: /Users/xinyu_peng/workspace/cvpr-2026/CVPR2026-diffusion-vfi/rebuttal.tex
% ...existing code...

\paragraph{Skip-concatenate sampling}
To mitigate the error accumulation in AR inference, we introduce the \textit{skip-concatenate sampling} strategy that achieves error accumulation at a constant level, regardless of the duration of the generated video. Our key insight is that for a strongly conditioned task like VFI, generating a new segment without referring to previous context will not introduce significant incoherence, as the global structure of the video has already been determined by the strong input condition. Therefore, we can generate an independent segment to reset the generative state and suppress accumulated errors. As illustrated in Figure~\ref{fig:concatenate-vs-causal}, unlike previous works that adapt the causal auto-regressive order~\cite{huang2025self}, our method first generates a \textit{skip chunk} independent of the most recent context, breaking the dependency chain to prevent error propagation. It then generates a \textit{concatenate chunk} conditioned on both the previous skip chunk and the new skip chunk, bridging the temporal gap to ensure temporal coherence. By periodically resetting the generative context with an independent skip chunk, this strategy effectively curbs long-term error accumulation while maintaining temporal continuity.

\section{Experiments}
\subsection{Implementation Details}

\paragraph{DiT training} We train our DiT model on the large-scale LAVIB~\cite{stergiou2024lavib} dataset. The model is initialized from pretrained WAN2.1 T2V model and is trained for 16,000 steps with a total batch size of 256, using flow matching loss (Eq.~\ref{eq:fm-loss}) with timestep shift of 5~\cite{esser2024scaling}. We use a fixed spatio-temporal resolution of $60\times512\times512$ by applying random spatial cropping. The spatial tiling size is set to $256\times256$ with a stride of 192, and the temporal chunk length is 20 frames. The temporal downsample factors for generating LQ videos are uniformly sampled from 2 to 16. We use AdamW optimizer and a constant learning rate schedule of 5e-5 with a ramp-up of 0.5 million samples.

\paragraph{VAE training} 
We finetune the VAE decoder from Wan2.1~\cite{wan2025} on the LAVIB dataset, keeping its encoder frozen. Following~\cite{rombach2022high}, the VAE is trained with a composite loss function comprising an L1 loss, a perceptual LPIPS loss~\cite{zhang2018unreasonable}, an adversarial loss~\cite{isola2017image}, and a KL-divergence penalty to regularize the latent space. These components are weighted by 1.0, 1.0, 0.5, and 1.0e-6, respectively. The training runs for 10,000 steps with a batch size of 256 and a learning rate of 2e-5 for both the VAE and discriminator. The discriminator is activated after 5,000 training steps.  We use a fixed resolution of $256\times 256$ and a temporal sequence length of 17 frames. Same as training DiT, the temporal downsample factors for generating LQ videos are uniformly sampled from 2 to 16.

\paragraph{Inference} All the results are obtained using Euler ODE solver of 16 sampling steps with timestep shift of 8. To address the challenge of processing ultra-high-resolution (e.g., 4K) videos, we leverage \textit{Ulysses Sequence Parallelism (USP)}~\cite{jacobs2023deepspeed}, which distributes the computation of a single long token sequence across multiple GPUs. Experiments show that using USP on two 80GB GPUs is sufficient for performing inference on 4K resolution videos. In the experiments, we use 4 GPUs for 4K resolution to further accelerate inference.

\paragraph{Datasets}
Our model operates on video chunks rather than adjacent frames, which makes direct comparison with prior works on conventional frame-centric benchmarks challenging, as these benchmarks typically only provide pairs of adjacent frames for each test case~\cite{xue2019video, choi2020channel,sim2021xvfi}. To address this limitation, we introduce two new video-centric benchmarks: SNU-FILM-entire and X4K-entire based on SNU-FILM~\cite{choi2020channel} and X4K1000FPS~\cite{sim2021xvfi} datasets, respectively. Specifically, for SNU-FILM-entire benchmark, we downsample each test video from the SNU-FILM dataset for $4\times$, $8\times$, and $16\times$, to construct low-frame-rate input videos. For X4K-entire, we downsample each test video from the X4K1000FPS dataset only for $16\times$ to construct low-frame-rate input videos since we observed that $4\times$, $8\times$ is relatively easy for current methods. Importantly, evaluated methods are allowed to access \textit{entire low-frame-rate videos} on these benchmarks. This not only allows for a comprehensive evaluation of our model's capabilities but also aims to foster future VFI research that can exploit broader temporal contexts beyond immediately adjacent frames. 

\paragraph{Metrics}
We report four metrics that jointly assess per-frame fidelity and temporal coherence: LPIPS~\cite{zhang2018unreasonable},  FVD~\cite{unterthiner2018towards},  VFIPS~\cite{hou2022perceptual}, and  FloLPIPS~\cite{danier2022flolpips}. We prioritize these metrics over conventional PSNR/SSIM because pixel-wise metrics correlate poorly with human perception~\cite{zhang2018unreasonable}, and generative models often achieve worse scores despite being rated better by human observers~\cite{watson2022novel,jain2024video}.

\subsection{Comparison with prior work}
\paragraph{Baselines} We compare against representative and widely-adopted VFI methods that support \textit{arbitrary-timestep VFI}, including optical flow-based RIFE~\cite{huang2022rife}, AMT~\cite{li2023amt}, EMA-VFI~\cite{zhang2023emavfi}, BiM-VFI~\cite{seo2025bim}, and diffusion-based EDEN~\cite{zhang2025eden}. As these models accept only adjacent frames, we run them frame-by-frame over the entire sequence to obtain the high-frame-rate videos. For EDEN, we achieve VFI beyond $2\times$ via recursive inference.

\paragraph{Quantitative results.} As shown in Tables~\ref{tab:main} and \ref{tab:x4k}, LDF-VFI demonstrates superior performance across multiple challenging benchmarks. Notably, it achieves state-of-the-art FVD scores across all evaluated settings---including SNU-FILM ($4\times$, $8\times$, $16\times$) and X4K ($16\times$)---while maintaining highly competitive VFIPS and FloLPIPS scores. The strong performance on these motion-aware metrics underscores the effectiveness of our video-centric approach in modeling and synthesizing large, complex motions. Furthermore, our significant improvements in motion-aware metrics over EDEN, a frame-centric diffusion-based baseline, highlight the inherent advantages of a video-centric paradigm. Finally, our method remains highly competitive in terms of per-frame visual quality, achieving LPIPS scores comparable to those of leading baselines.

\paragraph{Qualitative results} Figure~\ref{fig:compare-baseline} provides a qualitative comparison of our method against other VFI techniques on the SNU-FILM benchmark for $16\times$ interpolation. In challenging scenes with large and complex motion—such as the rapid flapping of a bird's wings or the fast movement of human legs. Our LDF-VFI generates significantly more plausible and temporally consistent intermediate frames. For instance, while baseline methods produce blurry or distorted results for the moving objects, our method successfully synthesizes sharp and coherent frames. This demonstrates the effectiveness of our approach in capturing long-range motion dynamics and preserving visual quality, even at high interpolation rates.

\begin{table*}[!t]
    \caption{Ablation study on model components. We conduct study on auto-regressive (AR) order (``causal'': conventional causal sampling v.s. ``skip-concate'': proposed skip-concatenate sampling), and VAE type (``Uncond.'': original WAN2.1 VAE decoder v.s. ``Cond.'': proposed conditional VAE decoder).}
    \centering
    \setlength{\tabcolsep}{5pt}
    \begin{tabular}{cccccccccccc}
\toprule
% \hline
        \multirow{2}{*}{AR order} & 
        \multirow{2}{*}{VAE type} & 
        \multicolumn{2}{c}{SNU-FILM-4$\times$} & 
        \multicolumn{2}{c}{SNU-FILM-8$\times$} & 
        \multicolumn{2}{c}{SNU-FILM-16$\times$}  &
        \multicolumn{2}{c}{Averaged} \\
         \cmidrule(lr){3-4} \cmidrule(lr){5-6} \cmidrule(lr){7-8}  \cmidrule(lr){9-10}
        &  & LPIPS$\downarrow$ & FVD$\downarrow$ & LPIPS$\downarrow$ & FVD$\downarrow$ & LPIPS$\downarrow$ & FVD$\downarrow$ & LPIPS$\downarrow$ & FVD$\downarrow$  \\
\midrule
% \hline
    Causal       & Uncond. & 0.045 & 14.22 & 0.055 & 19.48 & 0.087 & 34.32 & 0.062 & 22.67 \\
    Skip-concate & Uncond. & 0.043 & 9.05 & 0.053 & \textbf{14.55} & 0.084 & 27.55 & 0.060 \scriptsize{\color{blue}($\downarrow$3.2\%)} & 17.05 \scriptsize{\color{blue}($\downarrow$24.8\%)} \\
    \textbf{Skip-concate} & \textbf{Cond.}   & \CC{15}\textbf{0.032} & \CC{15}\textbf{8.21} & \CC{15}\textbf{0.045} & \CC{15}15.01 & \CC{15}\textbf{0.078} & \CC{15}\textbf{26.26} & \CC{15}\textbf{0.051} \scriptsize{\color{blue}($\downarrow$17.8\%)} & \CC{15}\textbf{16.49} \scriptsize{\color{blue}($\downarrow$27.2\%)} \\
% \hline
\bottomrule
    \end{tabular}
    \label{tab:abla_main}
\end{table*}

\begin{figure}[!t]
    \centering
    \includegraphics[width=1\linewidth]{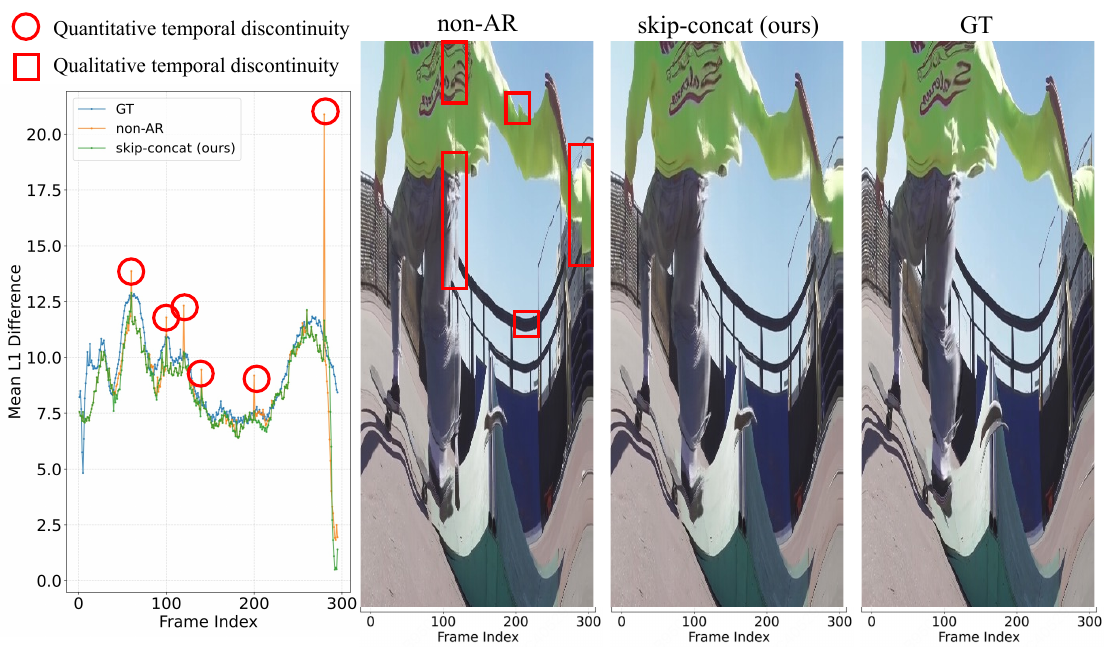}
    \vspace{-10pt}
    \caption{Temporal coherence analysis (non-AR vs. skip-concatenate AR). Spatio-temporal slices and mean $L_1$ pixel differences between adjacent frames show that our skip-concatenate approach resolves discontinuity artifacts at chunk boundaries.}
    \label{fig:non-AR}
\end{figure}

\begin{table}[!t]
    \caption{Inference efficiency comparisons. Running Time (RT) indicates the average generation time per frame on an NVIDIA H100 GPU ($\times4$ denotes USP on 4 GPUs).}
    \centering \small
    \setlength{\tabcolsep}{0.01pt}
    \begin{tabular}{c ccc ccc}
\toprule
        \multirow{2}{*}{Method} & \multicolumn{3}{c}{SNU-FILM-8$\times$} & \multicolumn{3}{c}{X4K-16$\times$} \\
        \cmidrule(lr){2-4} \cmidrule(lr){5-7} 
        & LPIPS$\downarrow$ & FVD$\downarrow$ & RT (s)$\downarrow$ & LPIPS$\downarrow$ & FVD$\downarrow$ & RT (s)$\downarrow$ \\
\midrule
        BiM-VFI        & 0.046 & 21.52 & 0.15 & 0.055 & 69.83 & 1.4 \\
        EDEN (2 steps) & 0.040 & 26.03 & 0.26 & 0.454 & 1586.99 & 4.8 \\
        \midrule
        LDF-VFI (2 steps)   & \CC{15}0.046 & \CC{15}21.25 & \CC{15}0.42 & \CC{15}0.087 & \CC{15}50.63 & \CC{15}2.0 $\times$4 \\
        LDF-VFI (4 steps)   & \CC{15}0.045 & \CC{15}19.00 & \CC{15}0.84 & \CC{15}0.074 & \CC{15}48.45 & \CC{15}4.0 $\times$4 \\
        LDF-VFI (8 steps)   & \CC{15}0.044 & \CC{15}16.85 & \CC{15}1.68 & \CC{15}0.074 & \CC{15}51.60 & \CC{15}8.1 $\times$4 \\
        LDF-VFI (16 steps)  & \CC{15}0.045 & \CC{15}15.01 & \CC{15}3.36 & \CC{15}0.071 & \CC{15}51.41 & \CC{15}16.3 $\times$4 \\
\bottomrule
    \end{tabular}
    \label{tab:abla_eff}
\end{table}

\begin{table}[!t]
    \caption{Comparative analysis of Full attention vs. Sparse attention. Evaluated at 6K training steps and 4 sampling steps.}
    \centering \small
    \setlength{\tabcolsep}{2pt}
    \begin{tabular}{c ccc ccc}
\toprule
        \multirow{2}{*}{Attention} & \multicolumn{3}{c}{SNU-FILM-8$\times$} & \multicolumn{3}{c}{X4K-16$\times$} \\
        \cmidrule(lr){2-4} \cmidrule(lr){5-7} 
        & LPIPS$\downarrow$ & FVD$\downarrow$ & RT (s)$\downarrow$ & LPIPS$\downarrow$ & FVD$\downarrow$ & RT (s)$\downarrow$ \\
\midrule
        Full  & 0.074 & 31.73 & 4.68 & 0.468 & 772.32 & 22.6 $\times$ 4 \\
        \textbf{Sparse}  & \CC{15}\textbf{0.070} & \CC{15}\textbf{24.98} & \CC{15}\textbf{3.36} & \CC{15}\textbf{0.157} & \CC{15}\textbf{120.40} & \CC{15}\textbf{4.0 $\times$ 4} \\
\bottomrule
    \end{tabular}  
    \label{tab:abla_attn}
\end{table}

\subsection{Ablation study}\label{sec:abla}

\paragraph{Effectiveness of skip-concatenate sampling}
Table~\ref{tab:abla_main} contrasts the conventional causal AR order with our proposed \emph{skip-concatenate} order. We observe that the skip-concatenate sampling strategy primarily improves temporal stability, yielding a significant reduction in FVD (e.g., from 22.67 to 17.05). This confirms that the proposed sampling order effectively mitigates temporal drift during auto-regressive generation. Furthermore, as shown in Figure~\ref{fig:non-AR}, we visualize spatio-temporal slices extracted along the time axis to compare with a non-AR baseline (using all skip chunks). Our AR approach with skip-concatenate effectively suppresses discontinuous jumps in temporal mean $L_1$ pixel differences between adjacent frames ($D_t = \|I_t - I_{t-1}\|_1$), avoiding the inconsistency artifacts observed in completely independent chunk generation.

\paragraph{Effectiveness of conditional VAE}
Table~\ref{tab:abla_main} shows that the conditional VAE further boosts visual quality over the unconditioned WAN2.1 decoder, primarily improving perceptual metrics like LPIPS. By injecting low-frame-rate (LQ) video embeddings throughout the multiscale features, the conditional VAE achieves sharper spatial fidelity. Together with the skip-concatenate sampling, it provides both superior temporal consistency and high per-frame quality.

\paragraph{Computational cost and attention mechanism} 
Table~\ref{tab:abla_eff} presents an inference efficiency comparison measured by the running time (RT) per generated frame on a single H100 GPU. By adjusting the sampling steps, LDF-VFI flexibly trades off visual quality for computational speed, remaining highly competitive even at just 2 steps. Furthermore, we examine the impact of full versus sparse attention in Table~\ref{tab:abla_attn}. For a fair comparison under high-resolution conditions, we analyze models with 6K training steps and 4 sampling steps. Full attention fails to generalize effectively to 4K resolution and requires significant longer RT. In contrast, our sparse attention dramatically reduces inference time and effectively generalize to 4K resolution, demonstrating its necessity for efficient scalability.

\begin{table}[!t]
	\caption{Key frame VAE reconstruction performance under different temporal upsampling on SNU-FILM-4$\times$ dataset. The results are averaged over 16 test videos of 17 frames.}
	\centering \small
	\setlength{\tabcolsep}{8pt}
	\begin{tabular}{lcccc}
\toprule
		 {Method} & PSNR$\uparrow$ & LPIPS$\downarrow$ & $L_1$$\downarrow$ \\
\midrule
		Zero padding & 28.40 & 0.011 & 0.004 \\
		\textbf{Nearest-neighbor}   & \CC{15}\textbf{30.87} & \CC{15}\textbf{0.006} & \CC{15}\textbf{0.002} \\
\bottomrule
	\end{tabular}
	\label{tab:abla_upsample}
\end{table}

\paragraph{Effectiveness of nearest-neighbor upsampling} Table~\ref{tab:abla_upsample} shows that the proposed nearest-neighbor upsampling strategy minimizes the reconstruction error of key frames and consistently outperforms conventional zero-padding upsampling across all evaluation metrics.

\section{Conclusion}

In this paper, we propose LDF-VFI, a holistic, video-centric framework that overcomes the fundamental limitations of conventional frame-centric VFI. Built upon an auto-regressive diffusion transformer, our approach leverages a novel skip-concatenate sampling strategy to effectively suppress error propagation, enabling temporally stable generation for videos of arbitrary length. Furthermore, we designed a conditional VAE decoder that incorporates LQ video information to further enhance reconstruction fidelity. Extensive evaluations on challenging video-centric benchmarks demonstrate that LDF-VFI achieves state-of-the-art performance. Our work marks a meaningful step toward exploring novel video-centric formulations for VFI.

% Acknowledgment

\section*{Acknowledgments}
This work was supported in part by the National Natural Science Foundation of China under Grant 62431017, Grant 62125109, Grant 62401357, Grant 62320106003, Grant U24A20251, Grant 62371288, in part by the National Key R\&D Program of China under Grant 2025YFF0515602, and in part by the Fundamental and Interdisciplinary Disciplines Breakthrough Plan of the Ministry of Education of China (JYB2025XDXM611).

% In this paper, we introduced LDF-VFI, a novel framework for video frame interpolation that addresses the critical challenges of temporal coherence and error accumulation in long video sequences. By adopting a holistic, video-centric modeling approach, our method moves beyond the limitations of traditional frame-pair-based techniques. The core of our framework is an auto-regressive diffusion transformer, which, combined with our proposed skip-concatenate sampling strategy, effectively mitigates error propagation during inference, enabling the generation of temporally stable videos of arbitrary length. Furthermore, our conditional VAE decoder enhances reconstruction fidelity by leveraging contextual information from the input video. Empirically,  LDF-VFI achieves state-of-the-art performance on challenging video-centric benchmarks, particularly in scenarios with large motion and high interpolation rates. Our work represents a significant step towards robust and scalable high-quality video frame interpolation, opening up new possibilities for applications requiring long-form video synthesis.

% \clearpage

{
    \small
    \bibliographystyle{ieeenat_fullname}
    \bibliography{main.bib}
}


\clearpage
\appendix
\setcounter{figure}{0}
\setcounter{table}{0}
\renewcommand{\thefigure}{A\arabic{figure}}
\renewcommand{\thetable}{A\arabic{table}}
\clearpage
\setcounter{page}{1}
\maketitlesupplementary
\setlength{\stripsep}{2pt plus 1pt minus 1pt}
\vspace*{-1.2em}

\begin{strip}
	\centering
	\refstepcounter{figure}\label{fig:vis}
	\includegraphics[width=1\textwidth]{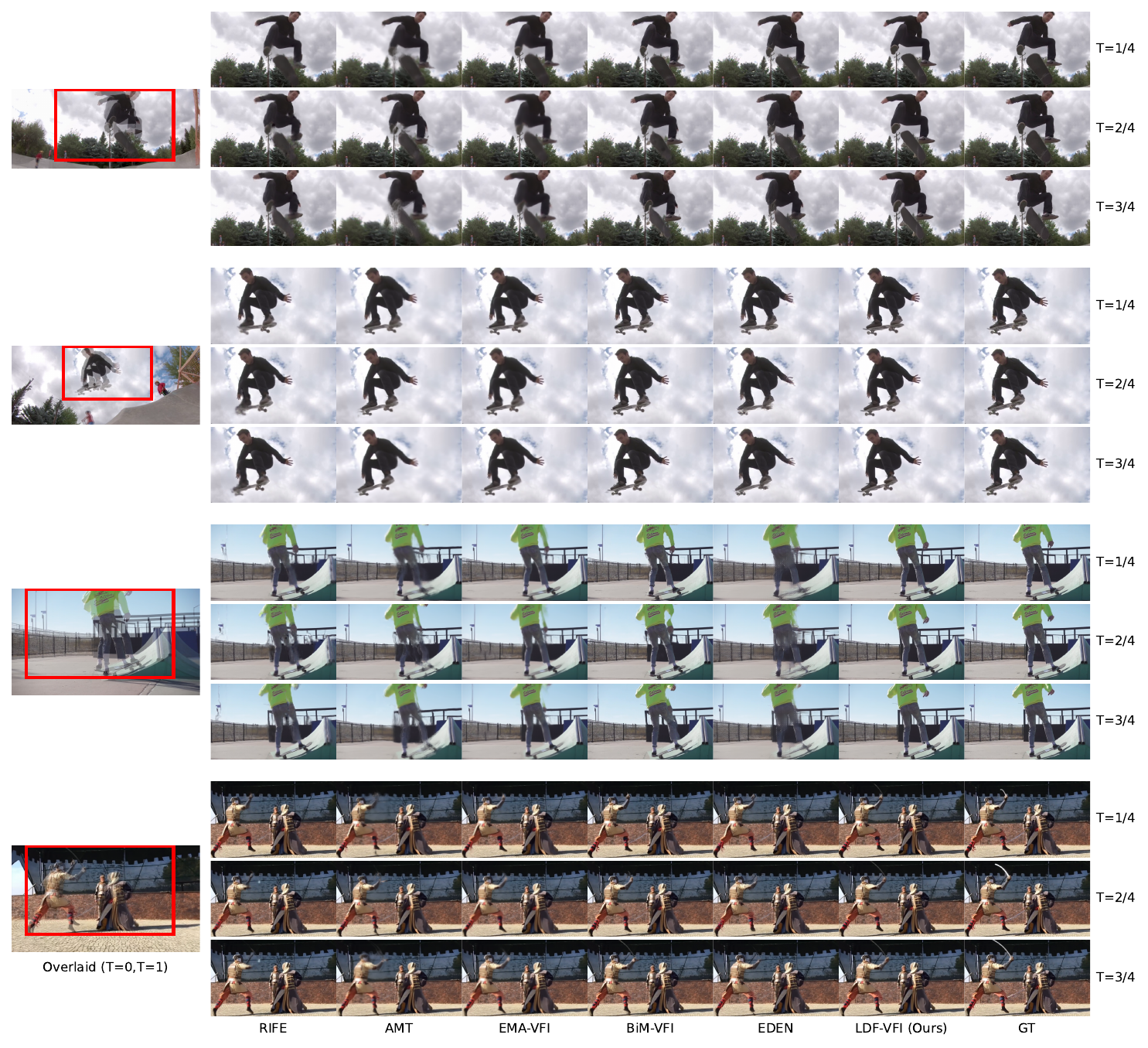}
	\vspace{0.2em}
	{\small\textbf{Figure \thefigure}. Additional visual comparisons.}
\end{strip}
        
\paragraph{More visualization} See Figure~\ref{fig:vis}.

\paragraph{Impact of temporal chunk length} Table~\ref{tab:abla_chunk} compares chunk sizes under similar training computation (32K steps for length 5 vs. 8K steps for length 20). Longer chunk improves temporal continuity, e.g., reducing X4K FVD from 510.63 to 121.29.

\begin{table}[!t]
	\caption{Impact of temporal chunk length under similar total computation.}
	\centering \scriptsize
	\setlength{\tabcolsep}{6pt}
	\renewcommand{\arraystretch}{0.92}
	\begin{tabular}{c cc cc}
\toprule
		\multirow{2}{*}{Temporal chunk} & \multicolumn{2}{c}{SNU-FILM-8$\times$} & \multicolumn{2}{c}{X4K-16$\times$} \\
		\cmidrule(lr){2-3} \cmidrule(lr){4-5}
		& LPIPS$\downarrow$ & FVD$\downarrow$ & LPIPS$\downarrow$ & FVD$\downarrow$ \\
\midrule
		Length 5  & \textbf{0.059} & 42.15 & 0.198 & 510.63  \\
		\textbf{Length 20} & \CC{15}0.066 & \CC{15}\textbf{24.87} & \CC{15}\textbf{0.126} & \CC{15}\textbf{121.29} \\
\bottomrule
	\end{tabular}
	\label{tab:abla_chunk}
\end{table}

\end{document}